# Early Detection of Parkinson's Disease using Motor symptoms and Machine Learning

Poojaa C[1], John Sahaya Rani Alex[1*]

[1] School of Electronics Engineering, Vellore Institute of Technology, Chennai, India

**ABSTRACT**

Parkinson's disease (PD) has been found to affect 1 out of every 1000 people, being more inclined towards the population above 60 years. Leveraging wearable-systems to find accurate biomarkers for diagnosis has become the need of the hour, especially for a neurodegenerative condition like Parkinson's. This work aims at focusing on early-occurring, common symptoms, such as motor and gait related parameters to arrive at a quantitative analysis on the feasibility of an economical and a robust wearable device. A subset of the Parkinson's Progression Markers Initiative (PPMI), PPMI Gait dataset has been utilised for feature-selection after a thorough analysis with various Machine Learning algorithms. Identified influential features has then been used to test real-time data for early detection of Parkinson Syndrome, with a model accuracy of 91.9%

## 1. INTRODUCTION

Parkinson's disease, a fast-growing condition, in elderly people, with the defining criteria including but not limited to rest tremor, rigidity, and impaired gait mostly along with the presence of bradykinesia. However, the clinical presentation is complicated and includes many non-motor symptoms as well [1]. Over 8.5 million people worldwide are estimated to have Parkinson's disease (PD) as of 2019. This global prevalence has doubled in the previous 25 years, making it one of the most common neurological disorders [2]. One of the most common and earliest symptoms is the motor symptoms in which people tend to develop a Parkinsonian Gait which is characterised by slow movement, bent posture, small and quick steps, and reduced swinging in their arms. Although these can be considered the pillars of diagnostic criteria, increasing gait difficulties and postural instability makes it hard to specify a distinguished biomarker [3].

The probable rate of misdiagnosis or an inaccurate diagnosis for a PD patient can be as high as 20%, especially if the diagnosis is done by a non-specialist since there is no objective test for it yet [4]. This brings into question the clinical assessments which can be influenced by various factors. In this case, wearable sensors, medical decision support tools, and research on specific neurodegenerative biomarkers can prove to be extremely useful and help in early diagnosis. With consumers embracing wearable technology, data acquired from these devices can be developed efficiently to be used as a secondary diagnostic tool. Moreover, automated analytics, when carried out effectively, can help physicians intervene sooner to diagnose the patient [5].

In this work, extensive Data Analysis was conducted on the Parkinson's Progression Marker Initiative (PPMI) Data, which was then used to extract high-accuracy feature sets. Furthermore, an inexpensive but efficient wearable device system has been designed to capture data from patients to predict whether they have PD.

## 2. REVIEW OF LITERATURE

In the past decade, the advancement of wearable sensors and devices has made it increasingly easier to adopt non-intrusive ways to monitor patients in controlled laboratory conditions as well as outside them. Additionally, the exponential growth of Machine Learning and Data Analytics techniques only makes it easier to develop robust and reliable systems for patients affected by various diseases. Significant work has been devoted to studying and detecting motor-movements and motion-related symptoms using wearable devices.

Asma Channa et al made a Systematic Review on the use of wearable devices in aiding the diagnosis, rehabilitation, assessment, and monitoring of patients with Parkinson's Disease or other neurocognitive disorders. The paper looks at 46 studies published between 2009 and 2020 to analyse how wearable devices can help improve the quality of life for these patients [6]. Luca Lonini et al. used six flexible sensors to detect tremors and bradykinesia in PD patients. Based on data from tasks completed, trained CNNs are deployed to find which movement exhibits signs of bradykinesia or tremor [7].

Bohan Shi et al used sensor-based technology to detect the freezing of gait (FOG) in the PD patient using Inertial motion sensors placed on the lower limbs of the neck and both ankles. A CNN model was used to identify FOG occurrences and the model achieved a geometric mean of 90.7% [8]. Aleksandr Talitckii et al. proposed a system based on wearable sensors wherein Sensor nodes were fixed on the dorsal part of the dominant hand and artificial intelligence was used to differentiate PD patients from those with similar diseases. Various machine-learning methods to analyse and classify tremors and bradykinesia with a final f1 score of 0.88 [9]. Nader Naghavi and Eric Wade, discuss the challenges in detecting the freezing of gait (FoG) in Parkinson's disease affected people and deploy transfer learning with a Deep Gait Anomaly Detector (DGAD) algorithm to improve FoG detection accuracy. With a simple walking task and sensors on the ankle, the effect of data augmentation and additional pre-FoG segments on prediction rates were evaluated in PD patients [10]. A neural network multilayer perceptron (MLP) model was proposed by He Juanjuan and co-authors for detecting bradykinesia in Parkinson's disease automatically. With two inertial sensors placed on the wrist and fingers, the system showed an accuracy of 90% for classifying PD patients and normal subjects and 85% accuracy for Parkinson's disease and Parkinson's syndrome [11]. Accelerometer readings were studied in the wild, and features were learned in Rubén San-Segundo and other co-authors' work. Among the twelve systems that were analysed, Spectra trained CNN model after Tremor Spectrum Extraction clubbed with Multilayer Perceptron led to the highest performance [12].

A comprehensive outlook on motor and non-motor symptoms with various studies and relevant work was carried out by Hanbin Zhang, Aditya Singh Rathore, and other co-authors. An insight into research challenges currently faced and the research gaps that can be explored such as embedded AI were thoroughly discussed [13]. Gunjan Pahuja & T. N. Nagabhushan studied and reviewed various Machine Learning Algorithms, analysed K-Nearest Neighbour, and Multilayer Perceptron, and found that Artificial Neural Networks (ANN) with Lev-Enberg–Marquardt algorithm give the highest classification accuracy of 95.89% for voice dataset of PD Patients [14]. Diagnosis of PDA was carried out by systematic analysis of ML, with different datasets, ML algorithms, objectives, tools, subjects, and associated outcomes. This was studied and analysed from various papers and journals by Arti Rana, Ankur Dumka, and co-authors [15].

# 3. MATERIALS AND METHODS

This section discusses in detail about the experimental setup, data acquisition and pre-processing, gait parameters and the data analysis techniques that have been adopted.

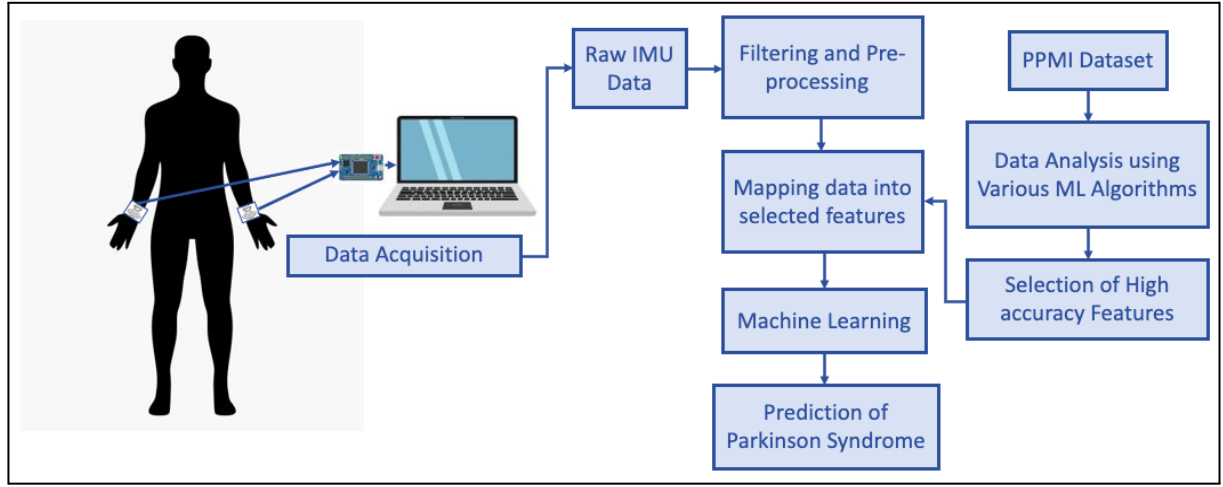

Fig. 1. Flow Diagram

## 3.1. DATASET - PPMI

The PPMI Data Repository is a repository containing clinical data collected through an ongoing longitudinal study consisting of capturing various kinds of data that show Parkinsonian symptoms, both motor and non-motor. Launched in 2010, the PPMI study assesses multiple large cohorts, and comprises evaluation of people with PD, to identify biomarkers of Parkinson Disease progression. PPMI aims at building the largest clinical, imaging, biologic-specimen dataset ever for Parkinson's community.

In order to obtain objective motor measurements that can be quantified so as to find pre-clinical symptoms or biomarkers was one of the reasons the gait study was proposed. A system with three-axial accelerometers, gyroscope and magnetometers have been used to form a sub-study within the PPMI genetic cohort to measure gait parameters, and has been validated on multiple large cohorts.

The PPMI dataset used in this work is a gait dataset that has motor features extracted from the raw accelerometer and gyroscope signals. A total of 81 patients were monitored for this particular dataset. The dataset has a sample size of 168, and contains 56 features, derived from six experimental tasks.

### 3.1.1. FEATURE SELECTION

After cleaning and pre-processing the dataset, few features were removed to prevent overfitting. A Mutual-Information algorithm was employed to check for high correlation between individual features and the predictor variable. The following results were obtained. Six of these high-accuracy features were selected in order to map the acquired data and for training the prediction model.

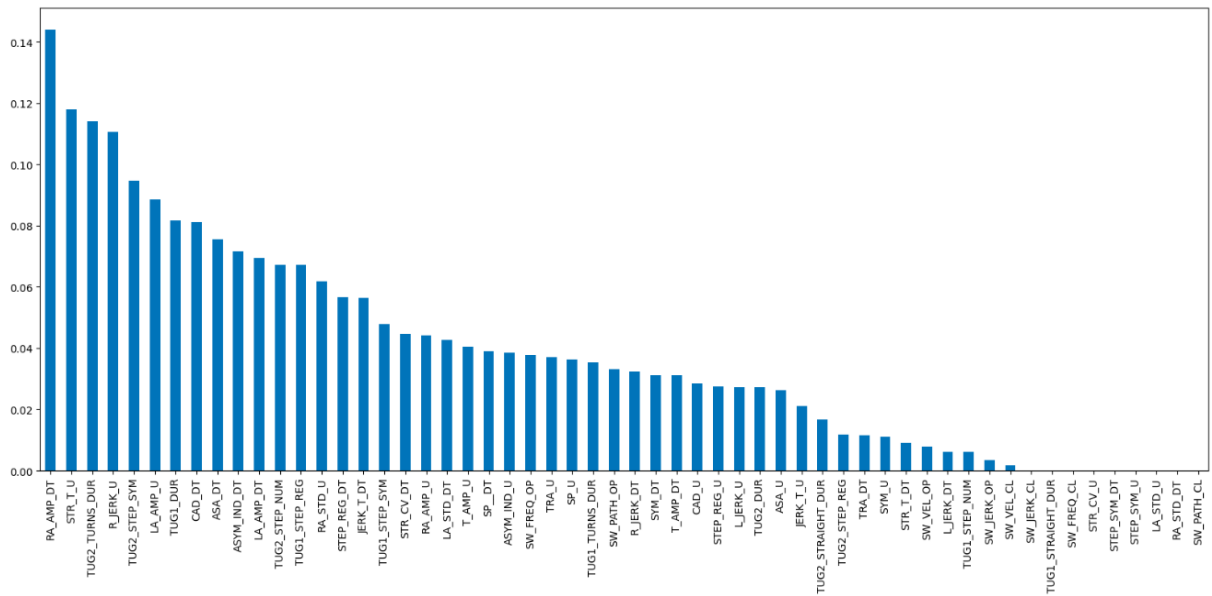

Fig. 2. Feature Selection using Mutual-Information

### 3.1.2. ML ALGORITHMS

After cleaning and preprocessing the PPMI dataset, various classification algorithms were used to analyse the PPMI dataset, as this would help provide a basis to select the appropriate ML algorithm to train and test the acquired data from the wearable system.

**SUPPORT VECTOR MACHINE**

Support Vector Machine, also known as SVM is a widely used supervised learning algorithm used for both regression and classification purposes, best suited for classification problems. It is known for its sound performance with just a limited no. of samples and its speed. Its mechanism involves classifying data points by means of a hyperplane in an N-dimensional space, wherein the dimensions of the hyperplane depends on the number of features.

**RANDOM FOREST**

Another supervised learning algorithm widely used in classification and regression tasks is Random Forest, which uses various samples to implement decision trees. In the case of classification, majority vote is taken whereas for regression, the average of their votes is taken. Its distinguished feature is that it can work on continuous variable dataset for regression tasks and also on datasets with categorical variables for classification.

**K-NEAREST NEIGHBOUR**

One of the most essential algorithms, K-Nearest Neighbour or KNN, is a non-parametric machine learning algorithm that has a number of applications such as data mining, pattern recognition, etc. It is used to make classifications or predictions based on proximity of the grouping of any single datapoint, Although it does not immediately learn while training, it stores the data, due to which prediction of a new unseen datapoint is easier, as it can be easily classified into a well suited category.

**LINEAR REGRESSION**

Linear Regression, an algorithm used for classification and predictive analysis. It finds the linear relationship between a dependent variable and an independent variable. It is a statistical model mostly used to predict a binary outcome.

**XGBOOST**

One of the most robust Machine Learning Algorithms, XGBoost, "Extreme Gradient Boosting", is known for its ability to be highly efficient and accurate, outperforming many other machine learning algorithms in tasks like ranking, regression and classification. It is used for scalable training, whilst being an ensemble learning method that makes a solid prediction from combining the predictions of multiple weak models. Its key highlights include its support for parallel-processing as well as the fact that it can handle data without significant pre-processing owing its ability to handle missing values in the dataset efficiently.

| Classifiers | Accuracy |
|---|---|
| Support Vector Machine | 66.0714 |
| Random Forest | 75 |
| K-Nearest Neighbour | 73.0303 |
| Logistic Regression | 64.2857 |
| XGBoost | 82.1429 |

Fig.3 Accuracy of ML Models on Gait analysis performed on the PPMI dataset

## 4. EXPERIMENTAL SETUP

All the experimental tasks carried out during data acquisition were adopted from the methods of PPMI data collection. Two tasks were selected, based on the feature selection, both the tasks majorly concentrating on features concerning arm-swing and walk:

1. **Task-1:** 1 minute normal walk in a straight line
2. **Task-2:** (Dual Task) 1 minute normal walk while serially subtracting a single digit number from a three-digit number throughout

These were chosen from the selection of feature sets from the data analysis of PPMI data.

### 4.1. DATA ACQUISITION

A cost-efficient, robust wearable-device system was designed for acquiring data from PD patients. The setup involves three Arduino Nano 33 BLE microcontroller boards which have inbuilt 9-axial Inertial Measurement sensors. Two of these microcontrollers serve as BLE Peripheral devices, which are placed at the back of the wrists and are responsible for transmitting IMU sensor data to the central device. The other one acts as a BLE Central device which receives the sensor data and captures it as a dataset (.csv file).

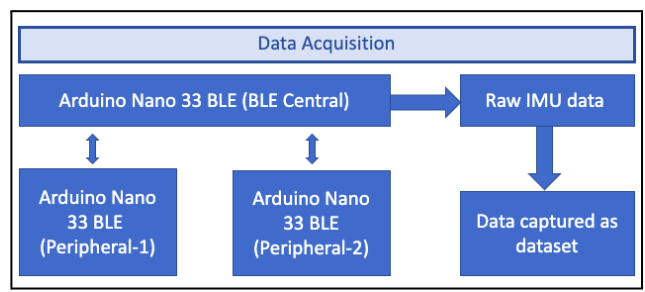

Fig.4. Gait Data Acquisition Flow

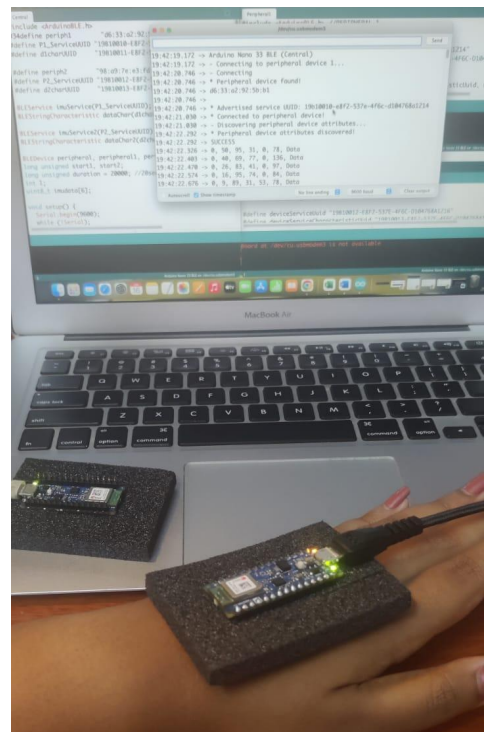

Fig.5. Real-time Data Acquisition

## 4.2. PREDICTION MODEL

After testing five ML algorithms, the K-Nearest-Neighbour classification algorithm was finalised for accurate prediction, and implemented on the selected features from the PPMI dataset. Although the initial accuracy of the model ranged around 70%, the model was hypertuned to get better results.

Finally, a test dataset, with no predictor variables, combined with the acquired dataset, after preprocessing, was used as test data on the model. The prediction accuracy of the model was found to be 91.99%.

## 5. RESULTS AND DISCUSSION

A cost-efficient wearable device system was designed to separately acquire PD patients' motor data. The raw is preprocessed, after which is used as test data in the KNN Model for prediction purposes. This robust mechanism would prove to be economical and useful in early detection, when monitored regularly.

On conducting thorough data analysis on the PPMI dataset, six of the dominating features were selected based on arm swing and walking tasks. These features were then tested on KNN-based prediction model to predict if the patient has Parkinson's.

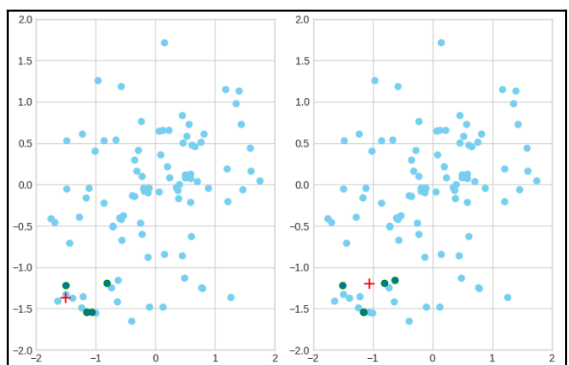

Fig. 6. Scatter plot of training data, test data on optimised KNN

Hypertuning the model with optimal parameters, along with using k-fold cross validation resulted in the model accuracy to 73%. However after testing the pre-trained model with the selected features, the prediction accuracy was found to be 91.99%

Quantitative approaches have shown potential in identifying small changes in those at risk for PD, whereas traditional tests used to detect motor impairments, such as the UPDRS, do not. The possibility of detecting pre-diagnosis motor alterations must therefore be increased by the use of more sensitive motor function testing. Gait and mobility quantitative measurements ought to offer a way to evaluate pre-diagnosis changes and gauge the development of the disease, which is something that can be potentially accomplished in the future.

Moreover, the search for accurate biomarkers for Parkinson's can be greatly influenced by the wearable device industry and the accuracy of automated analytical tools. This paves a path for further intensive Machine Learning and Deep Learning algorithms as well as for Embedded AI.

## 6. CONCLUSION

This work has examined the basic feasibility of designing a low-cost, robust wearable device for Parkinson's Disease, by implementing real-time data acquisition and early detection of PD. Various ML algorithms have been tested using the PPMI dataset and the KNN model has given the highest accuracy of 91.9%.